\documentclass[10pt,twocolumn,letterpaper]{article}

\usepackage{cvpr}              
\usepackage{algorithm,algorithmicx,algpseudocode}
\algrenewcommand{\algorithmiccomment}[1]{$\blacktriangleright$ #1}
\algnewcommand{\LineComment}[1]{\State \(\triangleright\) {\color{BlueViolet}#1}}

\usepackage{amsmath}
\usepackage{graphicx}
\usepackage{float}
\usepackage{multirow}
\usepackage{listings}
\usepackage{comment}
\usepackage[accsupp]{axessibility}
\usepackage[dvipsnames]{xcolor}

\definecolor{cvprblue}{rgb}{0.21,0.49,0.74}
\usepackage[pagebackref,breaklinks,colorlinks,citecolor=cvprblue]{hyperref}

\def\paperID{8028} 
\def\confName{CVPR}
\def\confYear{2024}

\title{UniPTS: A Unified Framework for Proficient Post-Training Sparsity}

\author{
Jingjing Xie$^{1}$, 
Yuxin Zhang$^{1}$, 
Mingbao Lin$^{2}$, 
Zhihang Lin$^{1}$, 
Liujuan Cao$^{1}$\thanks{Corresponding Author}, 
Rongrong Ji$^{1}$\\
$^{1}$Key Laboratory of Multimedia Trusted Perception and Efficient Computing,\\
Ministry of Education of China, School of Informatics, Xiamen University.\\
$^{2}$Tencent Youtu Lab\\
\tt\small \{jingjingxie, yuxinzhang, lmbxmu, zhihanglin\}@stu.xmu.edu.cn, \\
\tt\small \{caoliujuan, rrji\}@xmu.edu.cn
}

\let\oldequation\equation
\let\oldendequation\endequation

\renewenvironment{equation}
{\linenomathNonumbers\oldequation}
{\oldendequation\endlinenomath}

\begin{document}
\maketitle
\begin{abstract}
Post-training Sparsity (PTS) is a recently emerged avenue that chases efficient network sparsity with limited data in need.
Existing PTS methods, however, undergo significant performance degradation compared with traditional methods that retrain the sparse networks via the whole dataset, especially at high sparsity ratios.
In this paper, we attempt to reconcile this disparity by transposing three cardinal factors that profoundly alter the performance of conventional sparsity into the context of PTS.
Our endeavors particularly comprise (1) A base-decayed \textbf{sparsity objective} that promotes efficient knowledge transferring from dense network to the sparse counterpart. (2) A reducing-regrowing search algorithm designed to ascertain the optimal \textbf{sparsity distribution} while circumventing overfitting to the small calibration set in PTS. (3) The employment of dynamic sparse training predicated on the preceding aspects, aimed at comprehensively optimizing the \textbf{sparsity structure} while ensuring training stability.
Our proposed framework, termed UniPTS, is validated to be much superior to existing PTS methods across extensive benchmarks.
As an illustration, it amplifies the performance of POT, a recently proposed recipe, from 3.9\% to 68.6\% when pruning ResNet-50 at 90\% sparsity ratio on ImageNet.
We release the code of our paper at \url{https://github.com/xjjxmu/UniPTS}.

%
%

\end{abstract}

\section{Introduction}
\label{sec:intro}
\begin{figure*}[!htbp]
    \centering
    \includegraphics[width=0.9\textwidth,height=0.5\textheight, keepaspectratio]{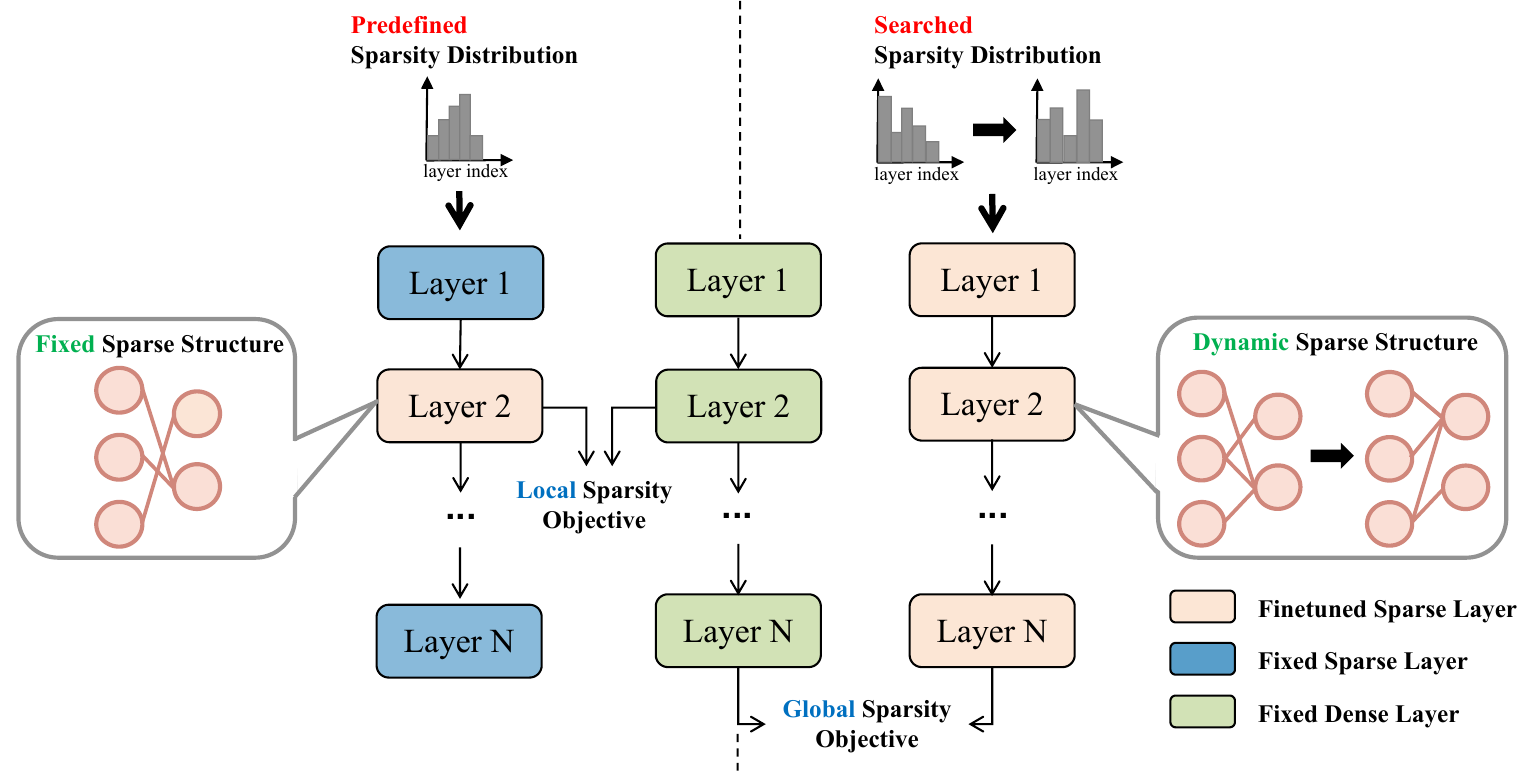}
    \caption{Comparison between POT and our UniPTS framework. \textbf{Left} shows that POT uses predefined sparsity distribution to obtain a fixed sparse structure and retrains the pruned layer with the local sparsity objective. But UniPTS(\textbf{Right}) searches the optimal sparsity distribution and leverages the global sparsity objective and dynamic sparsity training to explore optimal sparse structures.}
    \label{fig:framework}
    \vspace{-0.5\baselineskip}
\end{figure*}
Deep neural networks (DNNs) have shown exceptional performance in a variety of tasks, including computer vision~\cite{simonyan2014very,cai2019cascade,chen2017rethinking}, natural language processing~\cite{bert, improving},~\emph{etc}.
However, this extraordinary growth is offset by an overwhelming volume of model parameters, ranging from several millions to billions~\cite{brown2020language, xie2017aggregated}, presenting an impediment to DNN implementation in resource-constrained contexts. 
Accordingly, prodigious efforts have been channeled towards the evolution of model compression algorithms, encompassing model quantization~\cite{esser2019learned, nagel2020up}, network sparsity~\cite{han2015learning, evci2020rigging}, and knowledge distillation~\cite{hinton2015distilling, wu2022tinyvit}.


Notwithstanding the proficiency of these methods in shrinking the size of DNNs, they typically necessitate a model retraining phase via the full training set to recover the performance.
This could prove exceedingly burdensome in situational contexts marked by constraints both in terms of training resources and the accessibility of the dataset.
To circumvent this obstacle, researchers have developed post-training compression methodologies, which efficiently tune the compressed DNNs using a comparatively petite calibration dataset.
Most notable advancements to date have predominantly revolved around Post-Training Quantization (PTQ), where the quantized 8-bit model can reach performance on par with its full-precision counterpart~\cite{fang2020post}.
However, Post-Training Sparsity (PTS) has garnered relatively little attention, notwithstanding the commensurate prominence of sparsity in comparison with quantization for compressing DNNs.
One possible interpretation implicates that network sparsity methods rely more heavily on weight retraining,~\emph{w.r.t.}, iterative fine-tuning~\cite{han2015learning}, or even training from scratch~\cite{evci2020rigging}, to recover the performance. 
Thereby, considerable challenges emerge when merely minimal calibration datasets are accessible for such a retraining phase.
A recently proposed recipe, POT~\cite{lazarevich2021post}, sets a standard benchmark for PTS, undertaking a layer-wise minimization of the output discrepancy between sparse and dense weights via Mean Square Error (MSE) loss.
Despite its proficiency in preserving the performance at moderate pruning rates such as 50\%, POT encounters drastic performance loss at high sparsity rates, particularly deteriorating to the random level at 90\% sparsity, where traditional sparsity methods still manage to uphold performance.
Therefore, the argument over the necessity for data efficiency and maintenance of performance within network sparsity persists as an unresolved matter so far.
In this work, we present UniPTS as a practical remedy to ameliorate this issue.
As shown in \Cref{fig:framework}, UniPTS is a hybrid approach, meticulously crafted by investigating the failure of conventional sparsity methods in PTS scenarios from three vantage points including the \textbf{\textit{sparsity objective}}, \textbf{\textit{sparsity distribution}}, and \textbf{\textit{sparsity structure}}.
These aspects collaboratively contribute to the performance retention of sparse networks~\cite{evci2020rigging,lee2018snip,kusupati2020soft,frankle2018lottery,zhou2021effective}.
In particular, we first amend the \textbf{sparsity objective} from layerwise MSE~\cite{improving, li2021brecq, wei2022qdrop} in POT~\cite{lazarevich2021post} to a global Kullback-Leibler divergence, with its $log$ base adaptively evolves throughout the training schedule.
This not only fosters training acceleration, but also augments supervision from dense networks to sparse counterparts in a fluid manner.
Subsequently, we propose a novel evolutionary search algorithm to optimize the \textbf{\textit{sparsity distribution}},~\emph{i.e.}, layerwise sparsity ratios in PTS.
The principle innovations fall into an excessive sparsity allocation mechanism and a noise-disturbed fitness evaluation, which guarantees a robust search for the optimal solution while avoiding over-fitting to the small calibration set. 
Under the constraints of the preceding sparsity objective and sparsity distribution, we concludingly adopt the concept of Dynamic Sparsity Training (DST)~\cite{evci2020rigging, liu2020dynamic, liu2021sparse} to comprehensively explore the \textbf{\textit{sparsity structure}} in PTS instead of retraining the pruned network in a fixed~ typology~\cite{frankle2018lottery,lee2018snip,wang2020picking}.
%
%
%

%
In an expansive series of experiments spanning diverse computer vision tasks, we substantiate the efficacy of our proposed UniPTS.
The empirical evidence underscores that our method exhibits enhanced applicability to PTS, materializing a marked augmentation in performance, particularly prominent at elevated sparsity rates.
To illustrate, it improves the performance of POT from a meager 3.9\% to a considerable 68.6\% for pruning ResNet-50 at 90\% sparsity ratio, even using less training time.
Our work provides fresh insights into boosting the performance of PTS and we hope to encourage more research in probing the advantages of network sparsity through a pragmatic perspective.

\section{Related Work}
\label{sec:related}

\subsection{Post-Training Model Compression}
In quantization, various techniques have been developed to reduce the training overhead and data requirement, known as PTQ~\cite{nagel2020up,fang2020post}. 
However, existing methods often involve time-consuming retraining processes when it comes to sparsity.  POT~\cite{lazarevich2021post} is proposed as a pioneering pipeline for PTS to address this issue. 
POT suggests an iterative pruning to obtain the sparse network. In order to mitigate bias introduced by pruning, it follows a quantization methodology to correct bias~\cite{nagel2019datafree} and then fine-tunes the weights via the reconstruction of the feature map layer by layer. 
Although POT offers an initial insight into PTS, its performance degrades a lot at a high sparsity rate. 
%

\subsection{Sparsity Distribution}
Given a global sparsity rate, how to obtain an optimal layer-wise sparsity is an essential problem. Existing solutions can be categorized into heuristic-based, optimization-based, and search-based methods. 
Heuristic methods take advantage of the characteristics of each layer. For instance, the Erdos-Renyi-Kernel (ERK) allocates per-layer sparsity based on the number of parameters~\cite{evci2020rigging}. Such techniques rely on empirical analysis and cannot guarantee the optimal solution. 
Optimization-based methods learn the sparsity distribution by optimizing a pruning threshold or learnable masks ~\cite{kusupati2020soft,ning2020dsa}. 
Search-based methods aim to discover the optimal sparsity distribution via reinforcement learning~\cite{he2018amc} or evolutionary algorithms~\cite{liu2019metapruning}. These methods iteratively explore different sparsity configurations and evaluate their performance. 
These two kind of methods integrate the task objective with sparsity distribution and provide a flexible framework for sparsity allocation. However, most of them rely on regularization to meet the requirement of the global sparsity and need to tune hyperparameters delicately. 
We design an efficient evolutionary search for the optimal sparsity distribution. It can meet the requirement about the global sparsity without tuning hyperparameters and alleviate overfitting in PTS.
\subsection{Dynamic Sparsity Training}
Traditional network sparsity involves two steps: pruning and fine-tuning. 
Static sparsity training means that the sparse structure of the network remains unchanged after pruning and only the preserved weights are fine-tuned. Representative works of this type of method include one-shot pruning~\cite{lee2018snip} and gradual pruning~\cite{han2015learning}. 
Dynamic sparsity training represents that model's weights are dynamically pruned and regrow during fine-tuning. Evci \emph{et al}.~\cite{evci2020rigging} suggested the magnitude as the criterion for pruning and the gradient as the criterion for regrowth. 
In our paper, we find the traditional sparse training is not suitable for PTS, and modify both sparsity objective and training strategy.
%

\section{Method}
\begin{figure*}[!htbp]
    \centering
    \includegraphics[width=0.8\textwidth,height=0.8\textheight, keepaspectratio]{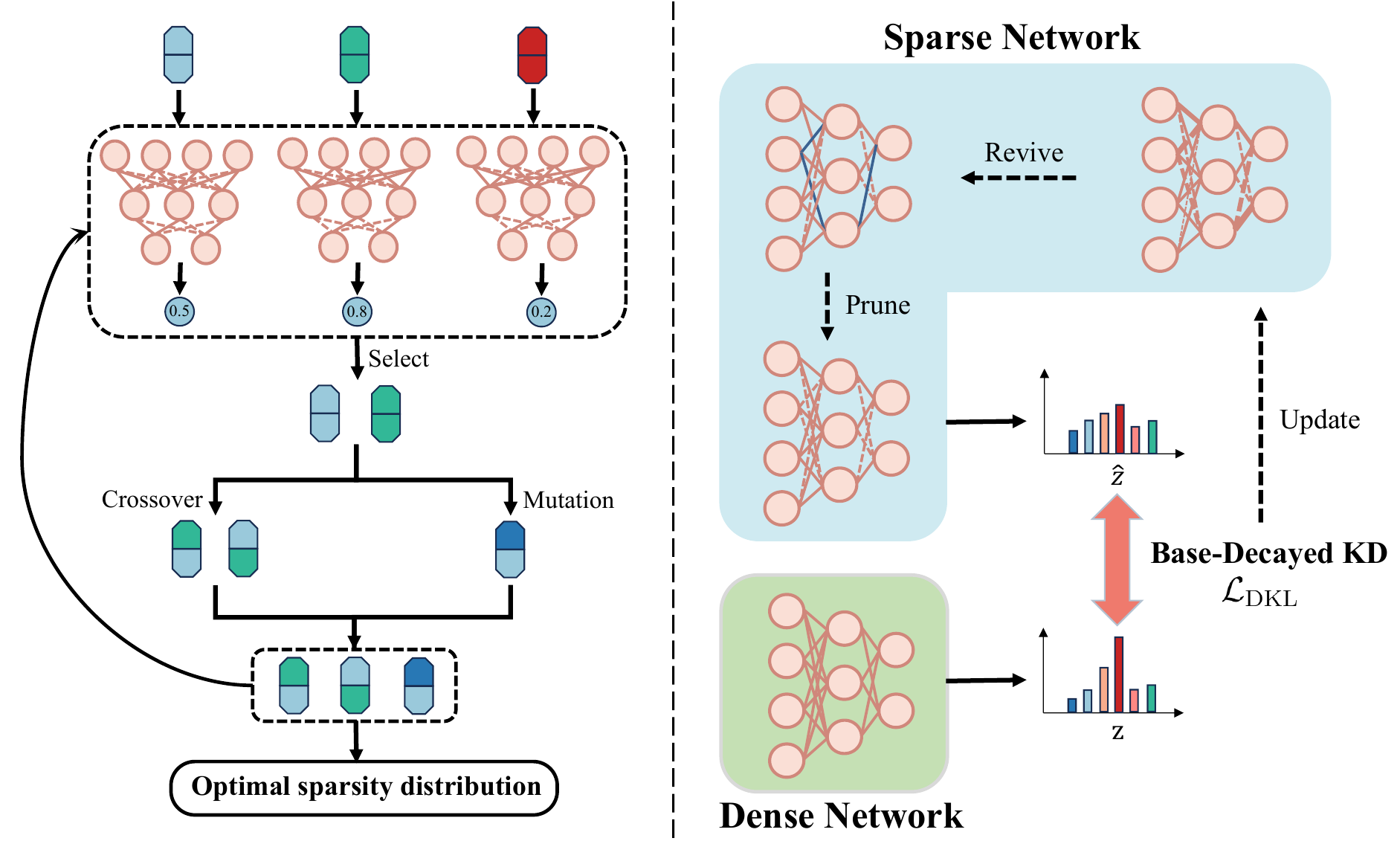}
    \caption{Overview about our method. \textbf{Left:}An overview of our method to search for sparsity distribution. We use evolutionary search to deal with the vast solution space. \textbf{Right:}We provide intuition for the training process. After finding optimal sparsity distribution, we use base-decayed KD loss and dynamic sparse training to retrain the pruned network.}
    \label{fig:pipeline}
    \vspace{-0.5\baselineskip}
\end{figure*}

\subsection{Background}
Primarily, we elucidate the fundamental concepts of network sparsity. 
Given the dense weights $\mathbf{W}$, network sparsity can be conceptualized as applying a binary mask $\mathbf{M}$ to $\mathbf{W}$.
A zero element within $\mathbf{M}$ signifies the removal of a specific weight, whereas non-zero elements indicate preservation.
The zero-masked weights alter network outputs,
%
which consequently results in negligible performance degradation in sparse networks, particularly at high sparsity rates~\cite{evci2020rigging,frankle2018lottery}.
To preserve the functionality of sparse networks, researchers have explored enhancing sparse networks from various perspectives including:
1) Employing training data to fine-tune the sparse network, with minimizing the discrepancy between the ground-truth labels and network outputs as the \textit{sparsity objective}~\cite{han2015learning}; 
2) Determining the \textit{sparsity distribution} across all layers, ~\emph{i.e.}, the relative sparsity ratio for each individual layer~\cite{kusupati2020soft,ning2020dsa,liu2019metapruning}; 
3) Exploring the \textit{sparsity structure} by modeling sparsity training to optimize the binary mask $\mathbf{M}$, therefore determining the removed weights for each layer~\cite{liu2020dynamic,liu2021sparse,zhou2021effective}. 

%

Albeit a plethora of methods proposed, 
they all necessitate substantial volumes of training data and computational costs. 
In scenarios characterized by limited resources, it becomes imperative to recover the performance of sparse networks solely through restrained quantities of data,
referred to as Post-Training Sparsity (PTS)~\cite{lazarevich2021post}. 
Regrettably, existing pruning methods demonstrate a deficit in scalability within post-training scenarios, enduring severe performance declines compared with full-data utilization.
Actually, the requirement for proficient and high-performing network sparsity reveals an unresolved dilemma in the current landscape.
\subsection{UniPTS}
%

In this section, we examine the underlying reasons for the failure of conventional sparsity methods in data-constrained scenarios and subsequently propose innovative UniPTS,  a unified framework to ameliorate their shortcomings and achieve proficient PTS.
Our efforts are articulated across the previously mentioned three dimensions -- \textit{sparsity objective}, \textit{sparsity distribution}, and \textit{sparsity structure} -- that are integral to recovering the performance of sparse networks. 

\subsubsection{Base-Decayed Sparsity Objective}
In traditional literature, the sparsity objective simply falls into fine-tuning the sparse network using training data, cost of which is basically the same as pre-training a dense network.
%
%
As a pioneering pipeline for PTS, POT~\cite{lazarevich2021post} utilizes the mean-squared error (MSE) loss as the training objective and fine-tunes per-layer weights independently:
\begin{equation}
    \mathcal{L}_{\text{MSE}} ={||\mathbf{Y}^{l}-\mathbf{\hat{Y}}^l||^{2}},
\end{equation}
where $\mathbf{Y}^l$ and $\mathbf{\hat{Y}}^l$ denote the $l$-th layer outputs of dense and sparse layers, respectively.
This supervision enables POT to effectively recover performance at moderate sparsity rates; however, its performance at high sparsity levels falls significantly behind traditional fine-tuning with full training data~\cite{zhou2021effective,tai2022spartan}.
%

%
%
%

The above layer-wise MSE is a localized metric that, if not well optimized, gradually gathers bias towards the final task prediction. An increasing sparsity rate challenges the match of individual values of $\mathbf{Y}^l$ and $\mathbf{\hat{Y}}^l$, and gives rise to expanded bias.
This analysis correlates with the phenomenon that POT drastically declines to randomized performance when confronted with a 90\% sparsity rate.

Given this, we choose to globally optimize the final prediction probability distributions $\mathbf{Z}$ from the dense output and $\mathbf{\hat{Z}}$ from the sparse output, and propose a base-decayed sparsity objective.
Kullback-Leibler divergence is used to quantify the difference between $\mathbf{Z}$ and $\mathbf{\hat{Z}}$:
\vspace{-0.5em}
\begin{equation}\label{eq:KL}
    \mathcal{L}_{\text{KL}}={\sum_{j=1}^C{\mathbf{Z}_{j}\log_{e}{\mathbf{Z}_{j}}/{\mathbf{\hat{Z}}_{j}}}},
    \smallskip
\end{equation}
where $C$ represents the total number of target classes.
Our principal impetus is to calibrate knowledge density transferred from the dense network to the sparse network, so as to avoid a sudden performance collapse at high sparsity rates.
From ~\cref{eq:KL}, it is apparent that the intensity of supervision correlates to the base of $log$ operation. 
We are deep in thought for the case of a high sparsity rate:
there has a substantial gap between the dense and sparse outputs in the early training stage, propelling training instability; as training proceeds, 
%
%
learning from the dense network results in a rapid decline of the KL loss, which however, brings about a very small gradient and further insufficient training.
Consequently, it is challenging to recover from the optimizing after-sparsified weights.
%
%
%
%
Therefore, we propose to decay the base of $log$ operation throughout the training process to adapt the loss scale as:
\vspace{-0.5em}
\begin{equation}\label{eq:dynamicloss}
    \mathcal{L}_{\text{DKL}}={\sum_{j=1}^C{\mathbf{Z}_{j}\log_{e \cdot \gamma^{t}}{\mathbf{Z}_{j}}/{\mathbf{\hat{Z}}_{j}}}},
\end{equation}
where $\gamma < 1$ represents the decay rate and $t$ is the current training epochs. 

As the base decreases, our base-decayed sparsity objective appropriately attenuates and amplifies loss scale during training. Therefore, it promotes an efficient knowledge transferring from dense to sparse networks.
It is important to highlight that our sparsity objective is significantly more efficient compared to the layer-wise MSE, as it fine-tunes the sparse networks in a global manner, eliminating the necessity for iterative retraining of each layer.

\subsubsection{Reducing-Regrowing Sparsity Distribution}

We proceed with the sparsity distribution issue.
Current cutting-edge methods mainly hinge on an automatic differentiable training~\cite{kusupati2020soft,zhou2021effective,zhang2022carrying}, which, however, becomes infeasible in the data-limited PTS.
Given this, we resort to exploiting evolutionary algorithms, efficacy of which has been well verified in structured network sparsity~\cite{liu2019metapruning,lin2020channel}.
%
%

Denote $\mathbf{W}=\{\mathbf{W}^{l}\}_{l=1}^L$ as $L$-layer pre-trained model with weights and $\mathbf{R}=\{r^l \}_{l=1}^L$ as the sparsity distribution candidate where $r^l$ signifies the sparsity rate of the $l$-th layer. 
Generally, as displayed in~\Cref{fig:pipeline}, an evolutionary algorithm generally initiates a set of individuals, each of which possesses a fitness score to measure the performance of its sparsity distribution.
The core of evolutionary algorithm is then to evolve these individuals from a series of crossover and mutation operations, and finally to pick up the one with best fitness.
Algorithm\,\ref{alg1} gives a comprehensive description of how to calculate the fitness.
Despite the success of these methods in structured sparsity~\cite{lin2020channel, li2020eagleeye}, a direct application in PTS encounters two dilemmas, as we analyze and address our uniques below.

\begin{algorithm}[!t]
\setlength{\abovedisplayskip}{0pt}
\setlength{\belowdisplayskip}{0pt}
    \caption{Fitness Calculation.}
    \label{alg1}
    \hspace*{0.02in} {\bf Input}:
    $L$-layer pre-trained model with weights $\mathbf{W} = \{\mathbf{W}^{l}\}_{l=1}^L$; Sparsity candidate $\mathbf{R} = \{r^l\}_{l=1}^L$.; Calibration set $\mathcal{D}$; Global sparsity rate $P$; Excessive sparsity rate $P_e$; \\
    \hspace*{0.02in} {\bf Output}: 
    Fitness of candidate $\mathbf{R}$ ; 
        \begin{algorithmic}[1]
            \LineComment{Model Sparsification.} 
            \State {Residual} = ($P_e$ $-$ $P$) $\times$ $numel$($\mathbf{W}$) \Comment{excessively sparsifying weights.}
	    \State $\mathbf{T}$ = \textrm{softmax}($\mathbf{R}$) $\times $ \textrm{Residual} \hspace{9.4mm}\Comment{regrow weights. }
             \State Sparsity = \{$P_e - \mathbf{T}^l/numel(\mathbf{W}^l)$\}
            \State $\hat{\mathbf{W}}$ = $Prune$($\mathbf{W}$, Sparsity)
            \vspace{1mm}
            \LineComment{Calibrate BN Statistics with Noisy Samples.}
            \State ${\mu},{\sigma}$ $\gets$ 0  \hspace{36.5mm}\Comment{reset statistics.}
            \For{Batch\_Data in $\mathcal{D}$}
                \State Noisy\_Batch\_Data = Batch\_Data + Gaussian\_Noise
                \State output = $\hat{\mathbf{W}}$(batch)
            \EndFor
            \vspace{1mm}
            \LineComment{Evaluate Calibration Accuracy as Fitness.} 
            \vspace{0.5mm}
            \State 
            \textrm{fitness} $\gets$ validate($\mathcal{D}$,$\hat{\mathbf{W}}$)
            \State \textbf{return} fitness;
	  \end{algorithmic} 
\end{algorithm}
First, colossal search space results from sparsifying weights, posing a challenge to ensure that the distribution candidate meets the desired global sparsity $P$.
To address this, we propose a reducing-and-regrowing sparsification method in Lines 2 -- 8 of Algorithm\,\ref{alg1} to obtain the sparse model $\hat{\mathbf{W}}$. 
Specifically, we introduce another sparsification $P_e > P$ at every layer so as to reduce the search space first.
Then, we further regrow the excessive sparsified weights using the evolutionary algorithm at a rate of $P_e - P$.

%
%
%
%
%

%
Second, assessing the fitness of a sparsity distribution candidate is also challenging, as evaluations easily lead to overfitting of the optimal structure to the calibration data,  raising a difficulty to assess the fitness of a distribution candidate.
Inspired by EagleEye~\cite{li2020eagleeye}, as shown in Lines 10--16 of Algorithm\,\ref{alg1}, for a sparse model with specific sparsity candidate, we recalculate the mean ${\mu}$ and variance ${\sigma}$ of batch normalization layers on the calibration set $\mathcal{D}$, and then choose the performance of the sparse network on the calibration set as the fitness vale of the sparsity candidate. To avoid the overfitting risk, we apply random Gaussian noise to the input samples prior to evaluating fitness.

\subsubsection{Sparsity Training}

%
%
%
%
%

%
The training of our UniPTS is to derive the final status of the binary mask $\mathbf{M}$ called sparsity structure under the constraints of the proposed: 1) base-decayed sparsity objective; 2) reducing-regrowing sparsity distribution.

Predominant DST methods leverage 
manually designated metrics like 
magnitude of weight gradients~\cite{evci2020rigging} to proceed weight pruning and regrowing at intervals.
However, within the scope of PTS, gradients lack reliability when determining essential weights, a consequence instigated by the limited quality and quantity of accessible data. 
We therefore consider the weight magnitude as the metric for pruning and regrowing.
Given weights of the $l$-th layer $\mathbf{W}^l$ and the corresponding sparsity rate $r^l$, the mask $\mathbf{M}^l$ during training can be derived as:
\begin{equation}
\mathbf{M}^{l}_{i,j}=\left\{
\begin{aligned}
1 &&& \text{if}\:|\mathbf{W}^{l}_{i,j}|
>\text{TopK}(|\mathbf{W}^{l}|, \left \lfloor (1-r^l)\times {S} \right \rfloor ), \\
0 &&& otherwise,\\
\end{aligned}
\right.
\end{equation}
where 
$\left |\cdot \right|$ is the absolute function and $\left \lfloor \cdot \right \rfloor$ is the floor operation.  $S$ = \text{numel($\mathbf{W}^l$)} is the number of parameters and $\text{TopK}(|\mathbf{W}^l|, \left \lfloor (1-r^l)\times {S} \right \rfloor )$ returns $\left \lfloor (1-r^l)\times {S} \right \rfloor$ largest value within $|\mathbf{W}^l|$.
Then, the forward output $\mathbf{\hat{Y}}^l$ is derived as:
%
%
\begin{equation}
     \mathbf{\hat{Y}}^l = ( \mathbf{W}^l \odot \mathbf{M}^l)\mathbf{\hat{Y}}^{l-1},
\end{equation}
where $\odot$ denotes the Hadamard product. 

Meanwhile, conventional DST methods commonly employ a periodic update sparsity structure, which means $\mathbf{M}$ will be updated every $\Delta T$ training iterations.
For PTS, we propose to use an iteration-wise sparse training strategy~\emph{i.e.}, $\Delta T=1$, as a periodic update may result in overfitting to a small amount of data. 
During training, we employ the Straight-Through Estimator ~\cite{bengio2013estimating} to approximate the gradient of pruned and unpruned weights during the backpropagation 
%
In this manner, the pruned weights are not reliant on the magnitude reduction of unpruned weights but can potentially recover through their own gradient, since both pruned and unpruned weights are subject to gradient updates.
%

Although such dynamic training enhances the diversity of sparse structures, it also aggravates fluctuation.
Following~\cite{zhou2021}, we modify the parameter update formula to prevent excessive fluctuation of the sparse structure and mitigate training instability issues. 
Specifically, a weight update mechanism is used to decay the magnitude of pruned weights by a certain proportion as:
\begin{equation}
\label{eq_decay}
\mathbf{W}_{i,j}^{l,t+1}=\left\{
\begin{aligned}
&\mathbf{W}_{i,j}^{l,t}-\beta*\dfrac{\partial \mathcal{L}}{\partial \mathbf{W}^{l,t}_{i,j}}, \quad \text{if}\:|\mathbf{W}_{i,j}^{l,t}| 
> \\&\quad\quad\quad\quad\quad\quad\,\,\,\text{TopK}(|\mathbf{W}^{l,t}|, \left \lfloor (1-r^l)\times {S} \right \rfloor ),\\ 
&\mathbf{W}_{i,j}^{l,t}-\beta*\dfrac{\partial \mathcal{L}}{\partial \mathbf{W}^{l,t}_{i,j}}-\alpha*\mathbf{W}_{i,j}^{l,t},\quad otherwise,\\
\end{aligned}
\right.
\end{equation}
where $t$ denotes iterations, $\beta$ represents the learning rate, and $\alpha$ is the decay proportion.  This adjustment can play a role in controlling the variation of the sparse structure by limiting the magnitude of the pruned weight. Such that, our sparse training process for PTS can explore sufficient sparse structures while maintaining training stability.

\begin{table*}[htbp!]
\caption{Image classification results on ImageNet-1K.}
\vspace{-1em}
\label{tbl:ImageNet}
\centering
\setlength{\tabcolsep}{3mm}{
\begin{tabular}{lcccccc}
\toprule  
&        &        &  \multicolumn{4}{c}{Top-1 Accuracy (\%)}  \\	
\cmidrule{4-7}
Model     & Dense Top-1 Accuracy (\%)  &    Sparsity Rate (\%)  &  POT    & RigL & STR & UniPTS\\
\midrule
\multirow{5}{*}{ResNet-18} & \multirow{5}{*}{69.76}    &  50 & 69.22& 69.18& 33.70 &\textbf{69.30}              \\
                           &                        &  60 & 68.31&68.64 & 32.59 &\textbf{68.51}    \\
                           &                        &  70 & 65.91&67.46 & 30.52 &\textbf{68.01} \\
                           &                        &  80 & 56.21&65.26 & 28.58 &\textbf{66.35} \\
                           &                        &  90 & 14.15&58.17 & 25.88 &\textbf{61.47}   \\
\midrule
\multirow{5}{*}{ResNet-50} & \multirow{5}{*}{76.12}    &  50 & 75.69&73.92& 53.22 &\textbf{75.76}              \\
                           &                        &  60 & 74.36& 73.20& 50.10 &\textbf{75.37}    \\
                           &                        &  70 & 69.96&71.87& 49.22 &\textbf{74.73} \\
                           &                        &  80 & 48.04&69.33& 48.32 &\textbf{73.10} \\
                           &                        &  90 & 3.90&61.68& 30.24 &\textbf{68.60}   \\
\midrule
\multirow{5}{*}{MobileNet-V2} & \multirow{5}{*}{72.05}&  50 & 69.25&66.57& 30.96 &\textbf{69.80}              \\
                           &                        &  60 & 63.39&64.75 & 20.57 &\textbf{68.01}    \\
                           &                        &  70 & 47.07 & 60.72& 14.62 &\textbf{64.93} \\
                           &                        &  80 & 9.13& 52.19& 9.40 &\textbf{59.47} \\
                           &                        &  90 & 0.2 & 30.44& 5.32 &\textbf{42.46}   \\
\bottomrule
\end{tabular}
}
\vspace{-1\baselineskip}
\end{table*}

\section{Experiments}

\subsection{Settings}

\textbf{Datasets and networks}. For ease of comparison, our experiments follow POT~\cite{lazarevich2021post} which included image classification using CNNs on the ImageNet-1K~\cite{5206848} and object detection using Faster-RCNN~\cite{ren2015faster} and SSD~\cite{liu2016ssd} on PASCAL VOC~\cite{everingham2010pascal}.
For image classification, we engage 10240 images from the ImageNet-1K training set to train sparse ResNet-18, ResNet-50~\cite{he2016deep} and MobileNet-V2~\cite{sandler2018mobilenetv2}. 
For ResNet-18/50, the initial learning rate is $0.01$ and the weight decay is $1\times10^{-4}$. For MobileNet-V2, the initial learning rate is $0.05$ and the weight decay is $4\times10^{-5}$.
For object detection, we use VGGNet-16~\cite{vgg} as the backbone for Faster-RCNN and MobileNet-V1~\cite{howard2017mobilenets} for SSD. We randomly select 10240 images from the training set which is composed of VOC2007 training set and VOC2012 training set.
The stochastic gradient descent (SGD) optimizer is leveraged for training sparse networks with 16000 iterations.
Uniformly across all networks, we implement a batch size of 64 and a cosine learning rate routine~\cite{loshchilov2016sgdr}. 

 \textbf{Implementation details}. We implement UniPTS using Pytorch~\cite{pytorch2015} and all experiments are conducted on a single NVIDIA RTX 3090. 
In particular, for the base-decayed sparsity objective in~\cref{eq:dynamicloss}, we set the decay rate $\gamma$ to 0.99.
For the reducing-regrowing sparsity distribution, the excessive sparsity rate $P_e$ is set as $(P+5)\%$.
For sparsity training in~\cref{eq_decay}, the decay proportion $\alpha$ is set as $3\times10^{-5}$.

 \textbf{Baselines}. We evaluate our proposed UniPTS in comparison to the sole established PTS technique, POT~\cite{lazarevich2021post}. 
In addition, we also compare the results of RigL~\cite{evci2020rigging}, a classical sparse training method, and STR~\cite{kusupati2020soft}, a differentiable search method for sparsity distribution, in the context of post-training implementation, to provide a more exhaustive comparison.

\subsection{Main Results}
\textbf{Image classification}.
Firstly, we report the quantitative results of various methods for pruning networks on image classification task, as outlined in \Cref{tbl:ImageNet}. 
Comparatively, UniPTS consistently outperforms POT across all settings, with the accuracy gap widening as the sparsity rate increases.
For instance, UniPTS is capable of boosting the accuracy of POT from a mere 3.9\% to 68.60\%, and that of MobileNet-V2 from 0.2\% to 42.46\% at a sparsity rate of 90\%. 
Moreover, we contrast UniPTS with conventional sparsity methods on PTS scenario. 
Predicated upon our Post-training design, UniPTS consistently surpasses traditional methods across all sparsity rates. 
The poor performance of STR underscore the fact that traditional differentiable training for sparsity distribution is impractical in data-limited PTS scenarios, thereby supporting our design for reducing-regrowing sparsity distribution. 
Furthermore, results from RigL reveal the significance of dynamic sparse training for PTS at high sparsity rates, reinforcing our design for sparse training.

 \textbf{Object Detection.} Moving beyond fundamental image classification benchmarks, we exploit the generalization capacity of UniPTS within the object detection task.
~\Cref{tbl:detect} compares our proposed UniPTS to POT for pruning Faster-RCNN~\cite{ren2015faster} and SSD~\cite{liu2016ssd} on PASCAL VOC~\cite{everingham2010pascal} at 90\% sparsity.
Notably, UniPTS yields robust performance improvement of $3.3$ and $3.4$ mAP for pruning Faster-RCNN~\cite{ren2015faster} and SSD~\cite{liu2016ssd}, respectively.
Given these favorable outcomes, the robustness and efficacy of UniPTS in object detection tasks are incontrovertibly confirmed.
\vspace{-1mm}
\begin{table}[!htbp]
\caption{Object detection results at 90\% sparsity rate.}
\vspace{-0.8em}
\label{tbl:detect}
\centering
{
\begin{tabular}{lccc}
\toprule  
Model & Method & mAP\\
\midrule
Faster-RCNN & POT & 51.29   \\
Faster-RCNN & UniPTS &  \textbf{54.59} \\
\midrule
SSD & POT & 57.02 \\
SSD & UniPTS & \textbf{64.42} \\
\bottomrule
\end{tabular}
}
\vspace{-0.5\baselineskip}
\end{table}

 \textbf{N:M sparsity.} In light of the likely demand for practical acceleration, we also appraise performance on the recently developed N:M semi-structured sparsity~\cite{zhou2021, zhang2022learning}, which stipulates at most N non-zero components within M consecutive weights to achieve expeditious inference aided by the N:M sparse tensor core~\cite{nvidia2020a100}.
The comparisons between UniPTS and POT on 2:4, 4:8 and 2:8 sparsity patterns are depicted in \Cref{tbl:nm}. 
It is evident that UniPTS can adeptly adapt to structured N:M sparsity. Regardless of the sparsity pattern, our method persistently supersedes POT by a noticeable margin. 
\vspace{-0.5\baselineskip}
\begin{table}[htbp]
\caption{Comparison between POT and UniPTS for N:M sparsity.}
\label{tbl:nm}
\centering
\scalebox{0.92}
{
\begin{tabular}{lcccc}
\toprule  
&                &  \multicolumn{3}{c}{Sparse Pattern}  \\	
\cmidrule{3-5}
Model      &    Method  &  2:4   & 4:8 & 2:8   \\
\midrule
\multirow{2}{*}{ResNet-18}    &  POT  &   66.52  & 67.33 & 54.43          \\
                              &  UniPTS &   \textbf{67.86}  & \textbf{68.30} & \textbf{63.98}   \\
\midrule
\multirow{2}{*}{ResNet-50}    &  POT  &   73.64  & 74.53 & 47.24         \\
                              &  UniPTS &   \textbf{74.83}  & \textbf{75.14} & \textbf{71.46}     \\
\midrule
\multirow{2}{*}{MobileNet-V2} &  POT  & 67.02   &  67.53   &  9.02                   \\
                              &  UniPTS & \textbf{68.78}   &  \textbf{69.17}   & \textbf{58.36}  \\
\bottomrule
\end{tabular}}
\vspace{-0.5\baselineskip}
\end{table}

\textbf{Pruning efficiency.} Moreover, we assess the pruning efficacy contrasting our proposed UniPTS and POT, delineated in~\Cref{tbl:cost}.
Compared to POT, UniPTS obviously holds an absolute advantage in the trade-off between pruning speed and accuracy.
It is also worth noting that supplanting the search phase for sparsity distribution with the immediate application of the ERK budget~\cite{evci2020rigging} attenuates pruning time to an exponential magnitude. Though this precipitates a modest decline in accuracy, it engenders an auxiliary alternative for users navigating scenarios characterized by resource scarcity.

%
%
%

%
%
%
%

%
\begin{table}[htbp]
\caption{Time cost for pruning at 90\% sparsity on ImageNet-1K. UniPTS* use ERK instead of searching sparsity distribution. }
\vspace{-0.8em}
\label{tbl:cost}
\centering
\scalebox{0.92}
{
\begin{tabular}{lccc}
\toprule  
Model & \hspace{0.5em}Method & Time  & Top-1  \\ 
& & cost (min) & Accuracy(\%) \\
\midrule
\multirow{2}{*}[-1ex]{ResNet-18} & POT & 140 & 14.15 \\
                           & UniPTS* & 25 & 61.24\\ 
                           & UniPTS & 260 &  61.47\\ 
\midrule
\multirow{2}{*}[-1ex]{ResNet-50} & POT & 484 &  3.90\\
                           & UniPTS* & 58 & 66.97\\
                           & UniPTS &  317 & 68.60 \\
\midrule
\multirow{2}{*}[-1ex]{MobileNet-V2} & POT & 536  &0.2 \\
                             & UniPTS* & 96 &  40.59 \\
                             & UniPTS & 357 &  42.46\\
\bottomrule
\end{tabular}
}
\vspace{-0.5\baselineskip}
\end{table}
%

%
%

\subsection{Ablation Study}
In this section, we investigate the efficacy of each component in our method.
To better understand the impact of these components on the overall performance, we conduct ablation experiments by replacing each component individually and show the performance on ImageNet-1K.

\textbf{Sparsity objective}. 
We first investigate the effect of our proposed base-decayed sparsity objective. In~\Cref{tbl:dynamic}, we examine the performance under five object variants including: 
1) global MSE: instead of layer-wise MSE like POT, we try global MSE between $\mathbf{Z}$ and $\mathbf{\hat{Z}}$. 2) cross entory (CE): we use task relevant loss as sparsity objective and fine-tune the sparse network. 
3) normal KL: we calculate the loss based on~\cref{eq:KL}; 4) dynamic temperature: we introduce a dynamic temperature to smooth prediction probability; 5) dynamic base: we calculate the loss based on~\cref{eq:dynamicloss}; 
As can be observed, our proposed base-decayed sparsity objective far surpasses other variants.

\begin{table}[!htbp]
\caption{Effect of sparsity objective when pruning ResNet-50 on ImageNet-1K.}
\vspace{-0.8em}
\label{tbl:dynamic}
\centering
\begin{tabular}{lcc}
\toprule  
Model & Strategy & Top-1 Accuracy (\%) \\
\midrule
ResNet-50 & global MSE & 13.07 \\
ResNet-50 & CE & 38.99 \\
\midrule
ResNet-50 & normal KL & 64.47 \\
ResNet-50 & dynamic temperature & 64.54 \\
ResNet-50 & dynamic base & \textbf{68.64} \\
\bottomrule

\end{tabular}
\vspace{-0.5\baselineskip}
\end{table}
%

\begin{figure}
    \centering
    \includegraphics[width=0.5\textwidth]{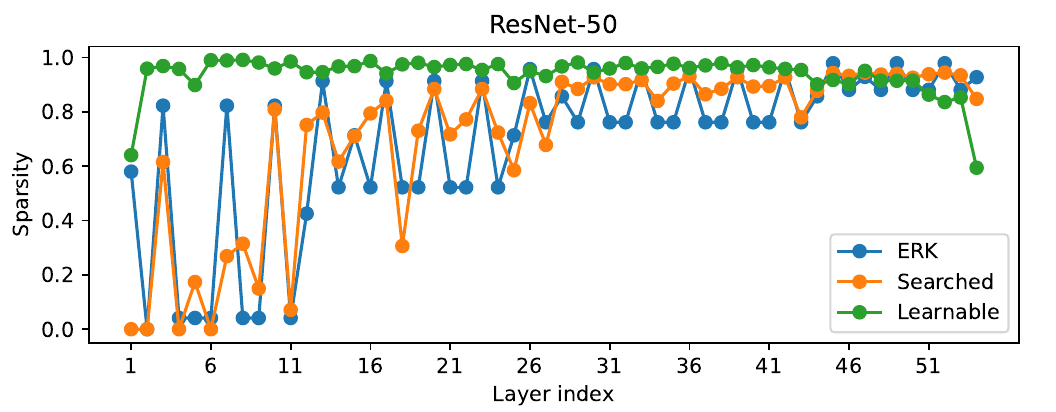}
    \vspace{-2em}
    \caption{Sparsity distribution obtained by different methods.}\label{fig:sparsity}
    \vspace{-1em}
\end{figure}

%
%
\textbf{Sparsity distribution}. To assess the effect of our proposed sparsity distribution search, we execute a comparative study involving alternative methods for determining the sparsity distribution. These include heuristic design based on ERK~\cite{evci2020rigging} and the learnable sparsity distribution derived from STR~\cite{kusupati2020soft}. 
~\Cref{tbl:sparsity} illustrates that our searched sparsity distribution yields the highest performance, with the margin of improvement escalating as the sparsity rate increases.
In addition, we extend this searched sparsity distribution to POT. As shown in ~\Cref{tbl:sparsity_pot}, the searched sparsity distribution can also contribute to improved performance in POT. As such, the effectiveness of our proposed sparsity distribution searching is validated.
We visualize the sparsity distribution across different layers obtained by each method, as shown in Fig.\,\ref{fig:sparsity}. Notably, the learnable method proved to be ineffective, as it merely maintained a nearly uniform sparsity rate across all layers. In contrast, UniPTS effectively identified elegant layer-wise sparsity rates (\emph{e.g.}, preserving more weights for the initial layers), thereby demonstrating its superior capability.

\begin{table}[!htbp]
\caption{Effect of sparsity distribution when pruning ResNet-50 on ImageNet-1K using UniPTS.}
\vspace{-0.8em}
\label{tbl:sparsity}
\centering
\scalebox{0.92}
{
\begin{tabular}{lccc}
\toprule  
Model & Method & Sparsity(\%)&Top-1 Accuracy (\%) \\
\midrule
ResNet-50 & ERK &90.00 & 66.97 \\
ResNet-50 & Learnable & 89.75  &  64.48 \\
ResNet-50 & Searched & 90.15 & \textbf{68.60} \\
\midrule
ResNet-50 & ERK &80.00 & 71.53 \\
ResNet-50 & Learnable & 79.94  &  71.86 \\
ResNet-50 & Searched &80.13 & \textbf{73.10} \\
\midrule
ResNet-50 & ERK &70.00 & 73.44 \\
ResNet-50 & Learnable & 69.95  &  74.19 \\
ResNet-50 & Searched &70.17 & \textbf{74.74} \\
\bottomrule
\end{tabular}
}
\vspace{-1\baselineskip}
\end{table}

\begin{table}[!htbp]
\caption{Effect of sparsity distribution when pruning ResNet-50 on ImageNet-1K using POT.}
\vspace{-0.8em}
\label{tbl:sparsity_pot}
\centering
\scalebox{0.92}
{
\begin{tabular}{lccc}
\toprule  
Model & Method & Sparsity(\%)&Top-1 Accuracy (\%) \\
\midrule
ResNet-50 & L2-norm &90.00 & 3.88 \\
ResNet-50 & Searched & 90.15 & \textbf{5.07} \\
\midrule
ResNet-50 & L2-norm &80.00 & 48.07 \\
ResNet-50 & Searched &80.13 & \textbf{50.85} \\
\midrule
ResNet-50 & L2-norm &70.00 & 69.95 \\
ResNet-50 & Searched &70.17 & \textbf{70.92} \\
\bottomrule
\end{tabular}
}
\end{table}

\begin{figure}[!t]
    \centering
    \raggedright 
    \includegraphics[width=0.4\textwidth]{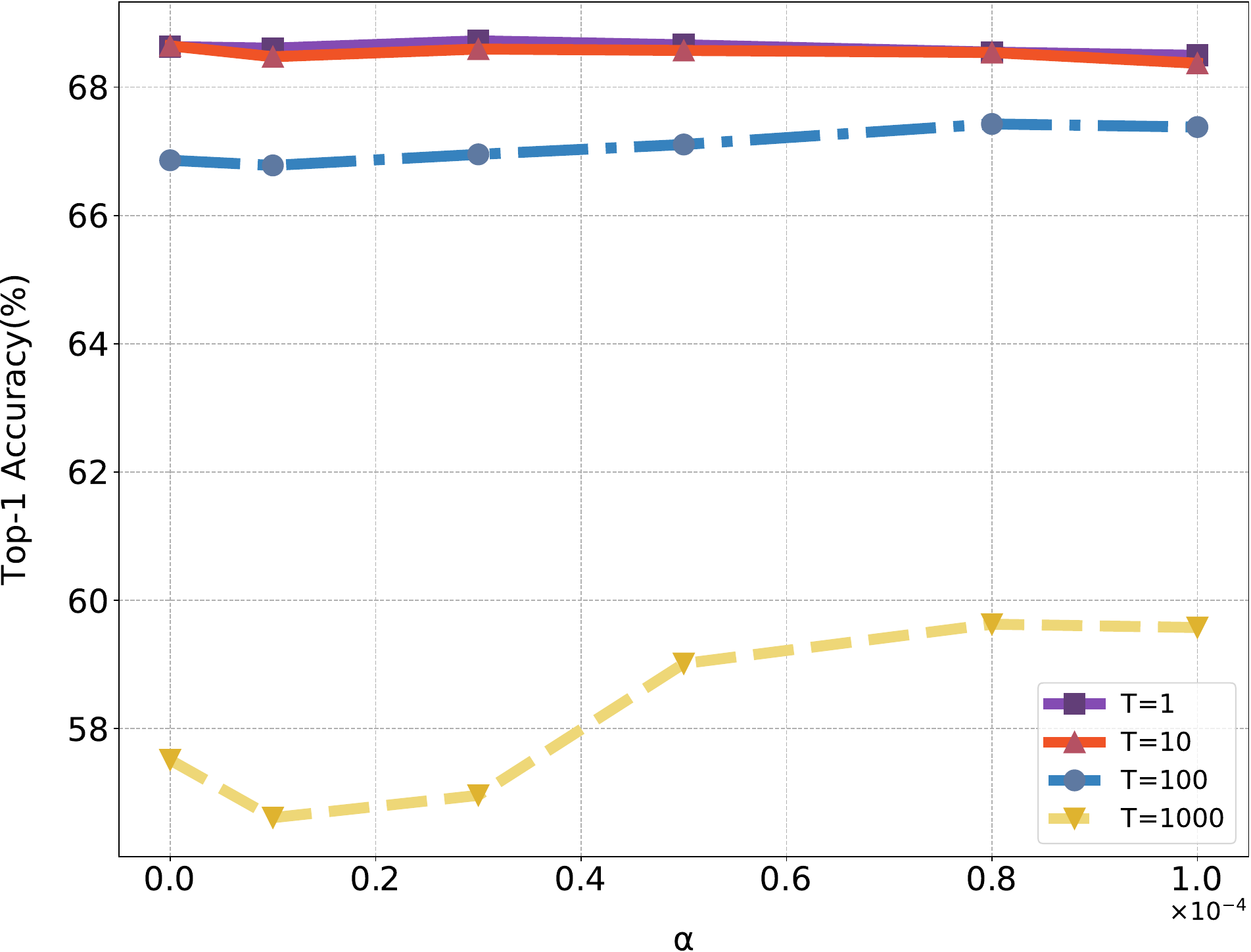}
    \caption{Performance influence of the sparse training strategy.}
    \label{fig:training}
\end{figure}

\textbf{Sparsity Training}. 
For sparsity training, we first investigate the effect of the decay factor $\alpha$ in~\cref{eq_decay}. 
~\Cref{fig:training} plots accuracy of pruned ResNet-50 at a sparsity rate of 90\% with different $\alpha$ adopted. 
It is intuitive that $\alpha$ indicates the degree of pruned weight decay, where $\alpha = 0$ indicates that pruned weights are updated exactly according to their gradient, and $\alpha \neq 0$ indicates pruned weights will decay to avoid reviving. 
With $\alpha$ increasing, the accuracy increases first and then decreases. A large $\alpha$ will lead to limited sparse structures. And a suitable $\alpha$ can constrain fluctuation and improve the performance.  

Furthermore, investigate the influence of the update interval $\Delta T$ for the pruning masks.~\Cref{fig:training} shows the performance of UniPTS under different intervals $\Delta T \in [1, 10, 100, 1000]$. 
As can be observed, the best accuracy is consistently obtained when the pruning mask is updated every iteration,~\emph{i.e.},~$\Delta T=1$.
Furthermore, as $\Delta T$ increases, performance gradually declines. For instance, the accuracy drops from 66.86\% to 57.51\% when $\alpha$ equals 0 and the interval increases from 100 to 1000.
This phenomenon appears to be in contrast with traditional practice in DST, where a larger interval yields better accuracy. 
Nonetheless, in the context of PTS, longer intervals imply that the learned mask will depend on the entirety of the validation set, leading to overfitting. 

%

%

\section{Conclusion}
In this paper, we focus on mitigating the performance gap between PTS and traditional network sparsity. 
We advance UniPTS, a unified framework comprising a base-decayed sparsity objective, a reducing-regrowing sparsity distribution, and a dynamic sparse training on the basis of the two preceding aspects to optimize the sparsity structure. Extensive experiments across a panoply of computer vision tasks demonstrate the effectiveness of UniPTS, which surpasses prior works by a significant margin, especially at high sparsity rates. Our work engenders new insights into enhancing the performance of PTS and we hope to stimulate further research into exploring the benefits of network sparsity from a more pragmatic perspective.
\section*{Acknowledgement}
This work was supported by National Science and Technology Major Project (No. 2022ZD0118202), the National Science Fund for Distinguished Young Scholars (No.62025603), the National Natural Science Foundation of China (No. U21B2037, No. U22B2051, No. 62176222, No. 62176223, No. 62176226, No. 62072386, No. 62072387, No. 62072389, No. 62002305 and No. 62272401), and the Natural Science Foundation of Fujian Province of China (No.2021J01002,  No.2022J06001).

%
%
%
%
%
{
    \small
    \bibliographystyle{ieeenat_fullname}
    \bibliography{main}

\begin{thebibliography}{46}
\providecommand{\natexlab}[1]{#1}
\providecommand{\url}[1]{\texttt{#1}}
\expandafter\ifx\csname urlstyle\endcsname\relax
  \providecommand{\doi}[1]{doi: #1}\else
  \providecommand{\doi}{doi: \begingroup \urlstyle{rm}\Url}\fi

\bibitem[Bengio et~al.(2013)Bengio, L{\'e}onard, and Courville]{bengio2013estimating}
Yoshua Bengio, Nicholas L{\'e}onard, and Aaron Courville.
\newblock Estimating or propagating gradients through stochastic neurons for conditional computation.
\newblock \emph{arXiv preprint arXiv:1308.3432}, 2013.

\bibitem[Brown et~al.(2020)Brown, Mann, Ryder, Subbiah, Kaplan, Dhariwal, Neelakantan, Shyam, Sastry, Askell, et~al.]{brown2020language}
Tom Brown, Benjamin Mann, Nick Ryder, Melanie Subbiah, Jared~D Kaplan, Prafulla Dhariwal, Arvind Neelakantan, Pranav Shyam, Girish Sastry, Amanda Askell, et~al.
\newblock Language models are few-shot learners.
\newblock \emph{Advances in Neural Information Processing Systems (NeurIPS)}, 33:\penalty0 1877--1901, 2020.

\bibitem[Cai and Vasconcelos(2019)]{cai2019cascade}
Zhaowei Cai and Nuno Vasconcelos.
\newblock Cascade r-cnn: High quality object detection and instance segmentation.
\newblock \emph{IEEE transactions on pattern analysis and machine intelligence (TPAMI)}, 43\penalty0 (5):\penalty0 1483--1498, 2019.

\bibitem[Chen et~al.(2017)Chen, Papandreou, Schroff, and Adam]{chen2017rethinking}
Liang-Chieh Chen, George Papandreou, Florian Schroff, and Hartwig Adam.
\newblock Rethinking atrous convolution for semantic image segmentation.
\newblock \emph{arXiv preprint arXiv:1706.05587}, 2017.

\bibitem[Deng et~al.(2009)Deng, Dong, Socher, Li, Li, and Fei-Fei]{5206848}
Jia Deng, Wei Dong, Richard Socher, Li-Jia Li, Kai Li, and Li Fei-Fei.
\newblock Imagenet: A large-scale hierarchical image database.
\newblock In \emph{IEEE Conference on Computer Vision and Pattern Recognition (CVPR)}, pages 248--255, 2009.

\bibitem[Devlin et~al.(2018)Devlin, Chang, Lee, and Toutanova]{bert}
Jacob Devlin, Ming-Wei Chang, Kenton Lee, and Kristina Toutanova.
\newblock Bert: Pre-training of deep bidirectional transformers for language understanding.
\newblock \emph{arXiv preprint arXiv:1810.04805}, 2018.

\bibitem[Esser et~al.(2019)Esser, McKinstry, Bablani, Appuswamy, and Modha]{esser2019learned}
Steven~K Esser, Jeffrey~L McKinstry, Deepika Bablani, Rathinakumar Appuswamy, and Dharmendra~S Modha.
\newblock Learned step size quantization.
\newblock \emph{arXiv preprint arXiv:1902.08153}, 2019.

\bibitem[Evci et~al.(2020)Evci, Gale, Menick, Castro, and Elsen]{evci2020rigging}
Utku Evci, Trevor Gale, Jacob Menick, Pablo~Samuel Castro, and Erich Elsen.
\newblock Rigging the lottery: Making all tickets winners.
\newblock In \emph{International Conference on Machine Learning (ICML)}, pages 2943--2952, 2020.

\bibitem[Everingham et~al.(2010)Everingham, Van~Gool, Williams, Winn, and Zisserman]{everingham2010pascal}
Mark Everingham, Luc Van~Gool, Christopher~KI Williams, John Winn, and Andrew Zisserman.
\newblock The pascal visual object classes (voc) challenge.
\newblock \emph{International Journal of Computer Vision (IJCV)}, 88:\penalty0 303--338, 2010.

\bibitem[Fang et~al.(2020)Fang, Shafiee, Abdel-Aziz, Thorsley, Georgiadis, and Hassoun]{fang2020post}
Jun Fang, Ali Shafiee, Hamzah Abdel-Aziz, David Thorsley, Georgios Georgiadis, and Joseph~H Hassoun.
\newblock Post-training piecewise linear quantization for deep neural networks.
\newblock In \emph{European Conference on Computer Vision (ECCV)}, pages 69--86, 2020.

\bibitem[Frankle and Carbin(2018)]{frankle2018lottery}
Jonathan Frankle and Michael Carbin.
\newblock The lottery ticket hypothesis: Finding sparse, trainable neural networks.
\newblock \emph{arXiv preprint arXiv:1803.03635}, 2018.

\bibitem[Han et~al.(2015)Han, Pool, Tran, and Dally]{han2015learning}
Song Han, Jeff Pool, John Tran, and William Dally.
\newblock Learning both weights and connections for efficient neural network.
\newblock \emph{Advances in Neural Information Processing Systems (NeurIPS)}, 28, 2015.

\bibitem[He et~al.(2016)He, Zhang, Ren, and Sun]{he2016deep}
Kaiming He, Xiangyu Zhang, Shaoqing Ren, and Jian Sun.
\newblock Deep residual learning for image recognition.
\newblock In \emph{IEEE Conference on Computer Vision and Pattern Recognition (CVPR)}, pages 770--778, 2016.

\bibitem[He et~al.(2018)He, Lin, Liu, Wang, Li, and Han]{he2018amc}
Yihui He, Ji Lin, Zhijian Liu, Hanrui Wang, Li-Jia Li, and Song Han.
\newblock Amc: Automl for model compression and acceleration on mobile devices.
\newblock In \emph{European Conference on Computer Vision (ECCV)}, pages 784--800, 2018.

\bibitem[Hinton et~al.(2015)Hinton, Vinyals, and Dean]{hinton2015distilling}
Geoffrey Hinton, Oriol Vinyals, and Jeff Dean.
\newblock Distilling the knowledge in a neural network.
\newblock \emph{arXiv preprint arXiv:1503.02531}, 2015.

\bibitem[Howard et~al.(2017)Howard, Zhu, Chen, Kalenichenko, Wang, Weyand, Andreetto, and Adam]{howard2017mobilenets}
Andrew~G Howard, Menglong Zhu, Bo Chen, Dmitry Kalenichenko, Weijun Wang, Tobias Weyand, Marco Andreetto, and Hartwig Adam.
\newblock Mobilenets: Efficient convolutional neural networks for mobile vision applications.
\newblock \emph{arXiv preprint arXiv:1704.04861}, 2017.

\bibitem[Kusupati et~al.(2020)Kusupati, Ramanujan, Somani, Wortsman, Jain, Kakade, and Farhadi]{kusupati2020soft}
Aditya Kusupati, Vivek Ramanujan, Raghav Somani, Mitchell Wortsman, Prateek Jain, Sham Kakade, and Ali Farhadi.
\newblock Soft threshold weight reparameterization for learnable sparsity.
\newblock In \emph{International Conference on Machine Learning (ICML)}, pages 5544--5555, 2020.

\bibitem[Lazarevich et~al.(2021)Lazarevich, Kozlov, and Malinin]{lazarevich2021post}
Ivan Lazarevich, Alexander Kozlov, and Nikita Malinin.
\newblock Post-training deep neural network pruning via layer-wise calibration.
\newblock In \emph{IEEE Conference on Computer Vision and Pattern Recognition (CVPR)}, pages 798--805, 2021.

\bibitem[Lee et~al.(2018)Lee, Ajanthan, and Torr]{lee2018snip}
Namhoon Lee, Thalaiyasingam Ajanthan, and Philip~HS Torr.
\newblock Snip: Single-shot network pruning based on connection sensitivity.
\newblock \emph{arXiv preprint arXiv:1810.02340}, 2018.

\bibitem[Li et~al.(2020)Li, Wu, Su, and Wang]{li2020eagleeye}
Bailin Li, Bowen Wu, Jiang Su, and Guangrun Wang.
\newblock Eagleeye: Fast sub-net evaluation for efficient neural network pruning.
\newblock In \emph{European Conference on Computer Vision (ECCV)}, pages 639--654, 2020.

\bibitem[Li et~al.(2021)Li, Gong, Tan, Yang, Hu, Zhang, Yu, Wang, and Gu]{li2021brecq}
Yuhang Li, Ruihao Gong, Xu Tan, Yang Yang, Peng Hu, Qi Zhang, Fengwei Yu, Wei Wang, and Shi Gu.
\newblock Brecq: Pushing the limit of post-training quantization by block reconstruction.
\newblock \emph{arXiv preprint arXiv:2102.05426}, 2021.

\bibitem[Lin et~al.(2020)Lin, Ji, Zhang, Zhang, Wu, and Tian]{lin2020channel}
Mingbao Lin, Rongrong Ji, Yuxin Zhang, Baochang Zhang, Yongjian Wu, and Yonghong Tian.
\newblock Channel pruning via automatic structure search.
\newblock \emph{arXiv preprint arXiv:2001.08565}, 2020.

\bibitem[Liu et~al.(2020)Liu, Xu, Shi, Cheung, and So]{liu2020dynamic}
Junjie Liu, Zhe Xu, Runbin Shi, Ray~CC Cheung, and Hayden~KH So.
\newblock Dynamic sparse training: Find efficient sparse network from scratch with trainable masked layers.
\newblock \emph{arXiv preprint arXiv:2005.06870}, 2020.

\bibitem[Liu et~al.(2021)Liu, Chen, Chen, Atashgahi, Yin, Kou, Shen, Pechenizkiy, Wang, and Mocanu]{liu2021sparse}
Shiwei Liu, Tianlong Chen, Xiaohan Chen, Zahra Atashgahi, Lu Yin, Huanyu Kou, Li Shen, Mykola Pechenizkiy, Zhangyang Wang, and Decebal~Constantin Mocanu.
\newblock Sparse training via boosting pruning plasticity with neuroregeneration.
\newblock \emph{Advances in Neural Information Processing Systems (NeurIPS)}, 34:\penalty0 9908--9922, 2021.

\bibitem[Liu et~al.(2016)Liu, Anguelov, Erhan, Szegedy, Reed, Fu, and Berg]{liu2016ssd}
Wei Liu, Dragomir Anguelov, Dumitru Erhan, Christian Szegedy, Scott Reed, Cheng-Yang Fu, and Alexander~C Berg.
\newblock Ssd: Single shot multibox detector.
\newblock In \emph{European Conference on Computer Vision (ECCV)}, pages 21--37, 2016.

\bibitem[Liu et~al.(2019)Liu, Mu, Zhang, Guo, Yang, Cheng, and Sun]{liu2019metapruning}
Zechun Liu, Haoyuan Mu, Xiangyu Zhang, Zichao Guo, Xin Yang, Kwang-Ting Cheng, and Jian Sun.
\newblock Metapruning: Meta learning for automatic neural network channel pruning.
\newblock In \emph{IEEE Conference on Computer Vision and Pattern Recognition (CVPR)}, pages 3296--3305, 2019.

\bibitem[Loshchilov and Hutter(2017)]{loshchilov2016sgdr}
Ilya Loshchilov and Frank Hutter.
\newblock Sgdr: Stochastic gradient descent with warm restarts.
\newblock In \emph{International Conference on Learning Representations (ICLR)}, 2017.

\bibitem[Nagel et~al.(2019)Nagel, van Baalen, Blankevoort, and Welling]{nagel2019datafree}
Markus Nagel, Mart van Baalen, Tijmen Blankevoort, and Max Welling.
\newblock Data-free quantization through weight equalization and bias correction, 2019.

\bibitem[Nagel et~al.(2020)Nagel, Amjad, Van~Baalen, Louizos, and Blankevoort]{nagel2020up}
Markus Nagel, Rana~Ali Amjad, Mart Van~Baalen, Christos Louizos, and Tijmen Blankevoort.
\newblock Up or down? adaptive rounding for post-training quantization.
\newblock In \emph{International Conference on Machine Learning (ICML)}, pages 7197--7206, 2020.

\bibitem[Ning et~al.(2020)Ning, Zhao, Li, Lei, Wang, and Yang]{ning2020dsa}
Xuefei Ning, Tianchen Zhao, Wenshuo Li, Peng Lei, Yu Wang, and Huazhong Yang.
\newblock Dsa: More efficient budgeted pruning via differentiable sparsity allocation.
\newblock In \emph{European Conference on Computer Vision (ECCV)}, pages 592--607, 2020.

\bibitem[Nvidia(2020)]{nvidia2020a100}
Nvidia.
\newblock Nvidia a100 tensor core gpu architecture.
\newblock \url{https://www.nvidia.com/content/dam/en- zz/Solutions/Data-Center/nvidia-ampere-architecture-whitepaper.pdf}, 2020.

\bibitem[Paszke et~al.(2019)Paszke, Gross, Massa, Lerer, Bradbury, Chanan, Killeen, Lin, Gimelshein, Antiga, et~al.]{pytorch2015}
Adam Paszke, Sam Gross, Francisco Massa, Adam Lerer, James Bradbury, Gregory Chanan, Trevor Killeen, Zeming Lin, Natalia Gimelshein, Luca Antiga, et~al.
\newblock Pytorch: An imperative style, high-performance deep learning library.
\newblock In \emph{Advances in Neural Information Processing Systems (NeurIPS)}, pages 8026--8037, 2019.

\bibitem[Radford et~al.(2018)Radford, Narasimhan, Salimans, Sutskever, et~al.]{improving}
Alec Radford, Karthik Narasimhan, Tim Salimans, Ilya Sutskever, et~al.
\newblock Improving language understanding by generative pre-training.
\newblock 2018.

\bibitem[Ren et~al.(2015)Ren, He, Girshick, and Sun]{ren2015faster}
Shaoqing Ren, Kaiming He, Ross Girshick, and Jian Sun.
\newblock Faster r-cnn: Towards real-time object detection with region proposal networks.
\newblock \emph{Advances in neural information processing systems}, 28, 2015.

\bibitem[Sandler et~al.(2018)Sandler, Howard, Zhu, Zhmoginov, and Chen]{sandler2018mobilenetv2}
Mark Sandler, Andrew Howard, Menglong Zhu, Andrey Zhmoginov, and Liang-Chieh Chen.
\newblock Mobilenetv2: Inverted residuals and linear bottlenecks.
\newblock In \emph{IEEE Conference on Computer Vision and Pattern Recognition (CVPR)}, pages 4510--4520, 2018.

\bibitem[Simonyan and Zisserman(2014)]{simonyan2014very}
Karen Simonyan and Andrew Zisserman.
\newblock Very deep convolutional networks for large-scale image recognition.
\newblock \emph{arXiv preprint arXiv:1409.1556}, 2014.

\bibitem[Simonyan and Zisserman(2015)]{vgg}
K. Simonyan and A. Zisserman.
\newblock Very deep convolutional networks for large-scale image recognition.
\newblock In \emph{International Conference on Machine Learning (ICML)}, 2015.

\bibitem[Tai et~al.(2022)Tai, Tian, and Lim]{tai2022spartan}
Kai~Sheng Tai, Taipeng Tian, and Ser~Nam Lim.
\newblock Spartan: Differentiable sparsity via regularized transportation.
\newblock \emph{Advances in Neural Information Processing Systems (NeurIPS)}, 35:\penalty0 4189--4202, 2022.

\bibitem[Wang et~al.(2020)Wang, Zhang, and Grosse]{wang2020picking}
Chaoqi Wang, Guodong Zhang, and Roger Grosse.
\newblock Picking winning tickets before training by preserving gradient flow.
\newblock \emph{arXiv preprint arXiv:2002.07376}, 2020.

\bibitem[Wei et~al.(2022)Wei, Gong, Li, Liu, and Yu]{wei2022qdrop}
Xiuying Wei, Ruihao Gong, Yuhang Li, Xianglong Liu, and Fengwei Yu.
\newblock Qdrop: Randomly dropping quantization for extremely low-bit post-training quantization.
\newblock \emph{arXiv preprint arXiv:2203.05740}, 2022.

\bibitem[Wu et~al.(2022)Wu, Zhang, Peng, Liu, Xiao, Fu, and Yuan]{wu2022tinyvit}
Kan Wu, Jinnian Zhang, Houwen Peng, Mengchen Liu, Bin Xiao, Jianlong Fu, and Lu Yuan.
\newblock Tinyvit: Fast pretraining distillation for small vision transformers.
\newblock In \emph{European Conference on Computer Vision (ECCV)}, pages 68--85, 2022.

\bibitem[Xie et~al.(2017)Xie, Girshick, Doll{\'a}r, Tu, and He]{xie2017aggregated}
Saining Xie, Ross Girshick, Piotr Doll{\'a}r, Zhuowen Tu, and Kaiming He.
\newblock Aggregated residual transformations for deep neural networks.
\newblock In \emph{IEEE Conference on Computer Vision and Pattern Recognition (CVPR)}, pages 1492--1500, 2017.

\bibitem[Zhang et~al.(2022{\natexlab{a}})Zhang, Lin, Lin, Chen, Wu, Tian, and Ji]{zhang2022carrying}
Yuxin Zhang, Mingbao Lin, Chia-Wen Lin, Jie Chen, Yongjian Wu, Yonghong Tian, and Rongrong Ji.
\newblock Carrying out cnn channel pruning in a white box.
\newblock \emph{IEEE Transactions on Neural Networks and Learning Systems (TNNLS)}, 2022{\natexlab{a}}.

\bibitem[Zhang et~al.(2022{\natexlab{b}})Zhang, Lin, Lin, Luo, Li, Chao, Wu, and Ji]{zhang2022learning}
Yuxin Zhang, Mingbao Lin, Zhihang Lin, Yiting Luo, Ke Li, Fei Chao, Yongjian Wu, and Rongrong Ji.
\newblock Learning best combination for efficient {N:}{M} sparsity.
\newblock In \emph{Advances in Neural Information Processing Systems (NeurIPS)}, 2022{\natexlab{b}}.

\bibitem[Zhou et~al.(2021{\natexlab{a}})Zhou, Yukun~Ma, Liu, Zhang, Yuan, Sun, and Li]{zhou2021}
Aojun Zhou, Junnan~Zhu Yukun~Ma, Jianbo Liu, Zhijie Zhang, Kun Yuan, Wenxiu Sun, and Hongsheng Li.
\newblock Learning n:m fine-grained structured sparse neural networks from scratch.
\newblock In \emph{International Conference on Learning Representations (ICLR)}, 2021{\natexlab{a}}.

\bibitem[Zhou et~al.(2021{\natexlab{b}})Zhou, Zhang, Xu, and Zhang]{zhou2021effective}
Xiao Zhou, Weizhong Zhang, Hang Xu, and Tong Zhang.
\newblock Effective sparsification of neural networks with global sparsity constraint.
\newblock In \emph{IEEE Conference on Computer Vision and Pattern Recognition (CVPR)}, pages 3599--3608, 2021{\natexlab{b}}.

\end{thebibliography}
}


\end{document}